# Learning Bayesian Network Parameters with Prior Knowledge about Context-Specific Qualitative Influences

**Ad Feelders** and **Linda C. van der Gaag**
Institute of Information and Computing Sciences, Utrecht University
PO Box 80089, 3508 TB Utrecht, The Netherlands

## Abstract

We present a method for learning the parameters of a Bayesian network with prior knowledge about the signs of influences between variables. Our method accommodates not just the standard signs, but provides for context-specific signs as well. We show how the various signs translate into order constraints on the network parameters and how isotonic regression can be used to compute order-constrained estimates from the available data. Our experimental results show that taking prior knowledge about the signs of influences into account leads to an improved fit of the true distribution, especially when only a small sample of data is available. Moreover, the computed estimates are guaranteed to be consistent with the specified signs, thereby resulting in a network that is more likely to be accepted by experts in its domain of application.

## 1 INTRODUCTION

For constructing a Bayesian network, often knowledge is acquired from experts in its domain of application. Experience shows that domain experts can quite easily and reliably specify the graphical structure of a network [7], yet tend to have more problems in coming up with the probabilities for its numerical part [8]. If data from every-day problem solving in the domain is available therefore, one would like to use these data for estimating the probabilities that are required for the graphical structure to arrive at a fully specified network. For many applications, unfortunately, the available data sample is quite small, giving rise to inaccurate estimates. The inaccuracies involved may then lead to a reasoning behaviour of the resulting network that violates common domain knowledge and the network will not easily be accepted by experts in the domain.

While domain experts often are found to have difficulties in coming up with probability assessments, evidence is building up that they feel more comfortable with providing qualitative knowledge about the probabilistic influences between the variables concerned [7, 11]. The qualitative knowledge provided by the domain experts, moreover, tends to be more robust than their numerical assessments. We demonstrated before that expert knowledge about the signs of influences between the variables in a Bayesian network can be used to improve the probability estimates obtained from small data samples [9]. We now extend our previous work to accommodate the wider range of context-specific signs, and context-specific independences more specifically. We argue that these signs impose order constraints on the probabilities required for the network. We then show that the problem of estimating probabilities under these order constraints is a special case of isotonic regression. Building upon this property, we present an estimator that is guaranteed to produce probability estimates that reflect the qualitative knowledge specified by the experts. The resulting network as a consequence is less likely to exhibit counterintuitive reasoning behaviour and is more likely to be accepted than a network with unconstrained estimates.

The paper is organised as follows. In the next section, we briefly review qualitative influences. In section 3, we discuss isotonic regression and provide an algorithm for its computation. We then show in section 4, that the problem of learning constrained network parameters is a special case of isotonic regression; we also discuss how the different constraints that result from qualitative influences are handled, and how the order constraints can be used in a Bayesian context. In section 5, we report on experiments that we performed on a small artificial Bayesian network and on a real-life network in the medical domain. Finally, we draw a number of conclusions from our work and indicate interesting directions for further research.

## 2 QUALITATIVE INFLUENCES

From a qualitative perspective, the variables in a Bayesian network may be related in different ways. In the sequel we assume all variables of the network to be binary. Let $\mathbf{X} = (X_1, \ldots, X_k)$ be the parents of a variable $Y$, and let $\Omega(\mathbf{X}) = \Omega(X_1) \times \Omega(X_2) \times \ldots \times \Omega(X_k) = \{0,1\}^k$ consist of vectors $\mathbf{x} = (x_1, x_2, \ldots, x_k)$ of values for the $k$ variables in $\mathbf{X}$, that is, $\Omega(\mathbf{X})$ is the set of all *parent configurations* of $Y$. Slightly abusing terminology, we sometimes say that $X_i$ *occurs* or *is present* if it has the value one. We write $\mathbf{X}_a$ for the sub-vector of $\mathbf{X}$ containing the variables $X_j$ for $j \in a$, where $a$ is a subset of $K = \{1, \ldots, k\}$. We further write $\mathbf{X}_{-a}$ for $\mathbf{X}_{K \setminus a}$.

A *qualitative influence* [14] between two variables expresses how observing a value for the one variable affects the probability distribution for the other variable. A *positive* influence of $X_i$ on $Y$ along an arc $X_i \to Y$ means that the occurrence of $X_i$ *increases* the probability that $Y$ occurs, regardless of any other direct influences on $Y$, that is, for all $\mathbf{x}, \mathbf{x}' \in \Omega(\mathbf{X})$ with $x_i = 1, x_i' = 0$ and $\mathbf{x}_{-i} = \mathbf{x}'_{-i}$, we have $p(y = 1|\mathbf{x}) \geq p(y = 1|\mathbf{x}')$. Similarly, there is a *negative* influence of $X_i$ on $Y$ along an arc $X_i \to Y$ if the occurrence of $X_i$ *decreases* the probability that $Y$ occurs, that is, for all $\mathbf{x}, \mathbf{x}' \in \Omega(\mathbf{X})$ with $x_i = 1, x_i' = 0$ and $\mathbf{x}_{-i} = \mathbf{x}'_{-i}$, we have $p(y = 1|\mathbf{x}) \leq p(y = 1|\mathbf{x}')$. A positive influence of $X_i$ on $Y$ is denoted by $X_i \xrightarrow{+} Y$ and a negative influence by $X_i \xrightarrow{-} Y$. An influence with either a positive or negative sign is called a *signed* influence. If no sign is specified, we call the influence *unsigned*.

The idea of signs of influences is readily extended to include context-specific signs [12]. A positive influence of $X_i$ on $Y$ within the context $\mathbf{X}_C = \mathbf{c}$, $C \subseteq \{1, \ldots, k\} \setminus i$, means that whenever $\mathbf{X}_C = \mathbf{c}$, the occurrence of $X_i$ increases the probability that $Y$ occurs, that is, for all $\mathbf{x}, \mathbf{x}' \in \Omega(\mathbf{X})$ with $\mathbf{x}_C = \mathbf{x}'_C = \mathbf{c}$, $x_i = 1, x_i' = 0$ and $\mathbf{x}_{-C \cup \{i\}} = \mathbf{x}'_{-C \cup \{i\}}$, we have $p(y = 1|\mathbf{x}) \geq p(y = 1|\mathbf{x}')$. A context-specific negative influence is defined analogously. A zero influence of $X_i$ on $Y$ within the context $\mathbf{X}_C = \mathbf{c}$ models a local context-specific independence (cf. [2]), that is, for all $\mathbf{x}, \mathbf{x}' \in \Omega(\mathbf{X})$ with $\mathbf{x}_C = \mathbf{x}'_C = \mathbf{c}$, $x_i = 1, x_i' = 0$ and $\mathbf{x}_{-C \cup \{i\}} = \mathbf{x}'_{-C \cup \{i\}}$, we have $p(y = 1|\mathbf{x}) = p(y = 1|\mathbf{x}')$. A signed context-specific influence of $X_i$ on $Y$ along an arc $X_i \to Y$ is denoted by $X_i \xrightarrow{s} Y[\mathbf{X}_C = \mathbf{c}]$, with $s \in \{+, -, 0\}$. Note that ordinary signed influences are special cases of context-specific influences with $C = \varnothing$. Further note that a signed influence in essence specifies a constraint on the parameters associated with a variable.

We assume, throughout the paper, that a domain expert provides the signs of the qualitative influences between the variables in a network. We would like to mention that for real-life applications these signs are quite readily obtained from experts by using a special-purpose elicitation technique tailored to the acquisition of signs of qualitative influences [11].

## 3 ISOTONIC REGRESSION

Our approach to obtaining parameter estimates for a Bayesian network that satisfy the signs of the influences specified by experts, is a special case of isotonic regression [13]. In this section we review isotonic regression in general; in the next section we discuss its application to parameter estimation for Bayesian networks.

Let $Z = \{z_1, z_2, \ldots, z_n\}$ be a nonempty finite set of constants and let $\preceq$ be a quasi-order on $Z$, that is:

1. for all $z \in Z$: $z \preceq z$ (reflexivity), and

2. for all $x, y, z \in Z : x \preceq y, y \preceq z \Rightarrow x \preceq z$ (transitivity).

Any real-valued function $f$ on $Z$ is *isotonic* with respect to $\preceq$ if, for any $z, z' \in Z$, $z \preceq z'$ implies $f(z) \leq f(z')$. We assume that each element $z_i$ of $Z$ is associated with a real number $g(z_i)$; these real numbers typically are estimates of the function values of an unknown isotonic function on $Z$. Furthermore, each element of $Z$ has associated a positive weight $w(z_i)$ that typically indicates the precision of this estimate. An isotonic function $g^*$ on $Z$ now is an *isotonic regression* of $g$ with respect to the weight function $w$ and the order $\preceq$ if and only if it minimizes the sum

$$\sum_{i=1}^{n} w(z_i) \left[f(z_i) - g(z_i)\right]^2$$

within the class of isotonic functions $f$ on $Z$. The existence of a unique $g^*$ has been proven by Brunk [5].

Isotonic regression provides a solution to many statistical estimation problems in which we have prior knowledge about the order of the parameters to be estimated. For example, suppose that we want to estimate binomial parameters

$$\mathbf{p} = (p(z_1), p(z_2), \ldots, p(z_n))$$

where $p(z_i)$ denotes the probability of success in population $z_i$. Let $n_i$ denote the number of observations sampled from population $z_i$, and let the number of successes $Y_i$ in this sample be binomially distributed with $Y_i \sim B(n_i, p(z_i))$. Then the isotonic regression of the estimates $\bar{Y}_i = Y_i/n_i$ with weights $w(z_i) = n_i$ provides the maximum-likelihood estimate of $\mathbf{p}$ given that $\mathbf{p}$ is isotonic on $(Z, \preceq)$. Note that this example suggests that the order-constrained estimates are obtained by first computing the unconstrained estimates and then performing the isotonic regression of these basic estimates with appropriate weights.

Isotonic regression problems can be solved by quadratic programming methods. Various dedicated algorithms, often restricted to a particular type of order, have been proposed as well. For $Z$ linearly ordered, for example, the *pool*

*adjacent violators* (PAV) algorithm is well-known. For our application, however, we require an algorithm that is applicable to sets of constants with arbitrary quasi-orders. For this purpose we will use the minimum lower sets (MLS) algorithm proposed by Brunk [3]. The MLS algorithm builds upon the concept of a lower set. A subset $L$ of $Z$ is a *lower set* of $Z$ if $z \in L$, $z' \in Z$, and $z' \preceq z$ imply $z' \in L$. The *weighted average* of a function $g$ on $Z$ for a nonempty subset $A$ is defined as

$$\text{Av}(A) = \frac{\sum_{z \in A} w(z) g(z)}{\sum_{z \in A} w(z)}$$

The algorithm now takes for its input the set of constants $Z = \{z_1, z_2, \ldots, z_n\}$ with quasi-order $\preceq$. With each $z_i \in Z$ again is associated a weight $w(z_i)$ and a real number $g(z_i)$. The algorithm returns the isotonic regression $g^*$ of $g$ with respect to $w$ and $\preceq$. The MLS algorithm resolves violations of the order constraints by averaging over suitably chosen subsets of $Z$. For the final solution, it partitions the set $Z$ into a number of subsets on which the isotonic regression is constant. The first subset $B_1$ in the final solution is a lower set of $(Z, \preceq)$. The second subset is a lower set of $(Z \setminus B_1, \preceq_2)$, where $\preceq_2$ is obtained from $\preceq$ by removing all order relations involving elements of $B_1$. This process is continued until $Z$ is exhausted. In each iteration the lower set with minimum weighted average is selected; in case multiple lower sets attain the same minimum, their union is taken.

    **MinimumLowerSets**($Z, \preceq, g(z), w(z)$)
    $\mathcal{L}$ = Collection of all lower sets of $Z$ wrt $\preceq$
    **Repeat**
        $B = \bigcup \{A \in \mathcal{L} \mid \text{Av}(A) = \min_{L \in \mathcal{L}} \text{Av}(L)\}$
        **For** each $z \in B$ **do**
            $g^*(z) = \text{Av}(B)$
        **For** each $L \in \mathcal{L}$ **do**
            $L = L \setminus B$
        $Z = Z \setminus B$
    **Until** $Z = \varnothing$
    **Return** $g^*$

The bottleneck of the algorithm from a computational point of view clearly is the generation of the lower sets, which is exponential in the size of the set of constants. We return to this observation in section 4.2.

## 4 PARAMETER LEARNING

In this section we address the maximum-likelihood estimation of parameters for a Bayesian network subject to the constraints imposed by the signs of influences, and show that it can be viewed as a special case of isotonic regression. We note that in the presence of signs, the parameters associated with the different parent configurations of a variable are no longer unrelated. Only those combinations of parameter values that are isotonic with respect to the quasi-order imposed by the specified signs, are feasible. The parameters associated with different variables are still unrelated however, because a sign imposes constraints on the parameters for a single variable only. We restrict our attention therefore to the parameters associated with a single variable.

### 4.1 ISOTONIC REGRESSION FORMULATION

To cast our problem of constrained parameter estimation into the isotonic regression framework, we proceed as follows. For parents $\mathbf{X}$ of $Y$, we construct an order $\preceq$ on $\Omega(\mathbf{X})$ in such a way that $\preceq$ corresponds to the order $\leq$ on the parameters $p(y = 1|\mathbf{x}), \mathbf{x} \in \Omega(\mathbf{X})$, that is implied by the specified signs. More specifically, for any qualitative influence $X_i \xrightarrow{s} Y[\mathbf{X}_C = \mathbf{c}], s \in \{+, -, 0\}$, we impose the following order on $\Omega(\mathbf{X})$: for all $\mathbf{x}, \mathbf{x}' \in \Omega(\mathbf{X})$ with $\mathbf{x}_C = \mathbf{x}'_C = \mathbf{c}$, $x_i = 1, x'_i = 0$ and $\mathbf{x}_{-C \cup \{i\}} = \mathbf{x}'_{-C \cup \{i\}}$:

- if $s = +$ then $\mathbf{x}' \preceq \mathbf{x}$, since the positive sign implies $p(y = 1|\mathbf{x}') \leq p(y = 1|\mathbf{x})$;

- if $s = -$ then $\mathbf{x} \preceq \mathbf{x}'$, since the negative sign implies $p(y = 1|\mathbf{x}) \leq p(y = 1|\mathbf{x}')$;

- if $s = 0$ then $\mathbf{x} \preceq \mathbf{x}'$ and $\mathbf{x}' \preceq \mathbf{x}$, since the zero enforces the equality $p(y = 1|\mathbf{x}') = p(y = 1|\mathbf{x})$.

The other ordering statements follow from the transitivity and reflexivity properties of quasi-orders. The specified influences constrain the parameters $p(y = 1|\mathbf{x})$ to be non-decreasing on $(\Omega(\mathbf{X}), \preceq)$.

Now suppose that we have available a data set $\mathcal{D}$ from which we would like to estimate the parameters $p(y = 1|\mathbf{x})$. The unconstrained maximum-likelihood estimate of $p(y = 1|\mathbf{x})$ is given by

$$\hat{p}(y = 1|\mathbf{x}) = \frac{n(y = 1, \mathbf{x})}{n(\mathbf{x})}$$

where $n(y = 1, \mathbf{x})$ denotes the number of observations in $\mathcal{D}$ with $y = 1$ and $\mathbf{X} = \mathbf{x}$.

The following observation now links isotonic regression to the problem currently considered: the isotonic regression $p^*(y = 1|\mathbf{x})$ of $\hat{p}(y = 1|\mathbf{x})$ with weights $w(\mathbf{x}) = n(\mathbf{x})$ provides the maximum-likelihood estimate of $p(y = 1|\mathbf{x})$, for all $\mathbf{x} \in \Omega(\mathbf{X})$, subject to the constraint that these estimates must be non-decreasing on $(\Omega(\mathbf{X}), \preceq)$ ([4], see also [13] page 32).

To illustrate the above observation, we consider a fragment of a Bayesian network. Let $\mathbf{X} = (X_1, X_2, X_3)$ be the parents of a variable $Y$, with qualitative influences on $Y$ as shown in figure 1. The fragment expresses the prior knowledge that $X_1$ has a positive influence on $Y$ and that, if $X_1$ is absent, $X_3$ has a negative influence on $Y$; it further models that if $X_1$ is present and $X_2$ is absent, then $X_3$ has no influence on $Y$. The network does not specify any prior knowledge about the sign of the influence of $X_2$ on $Y$.

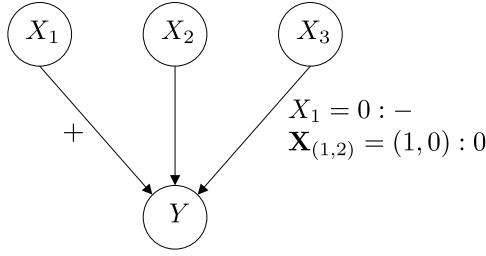

Figure 1: An example network fragment

Figure 2 shows the quasi-order on the parent configurations that is imposed by the specified influences, where an arrow from $\mathbf{x}$ to $\mathbf{x}'$ indicates that $\mathbf{x}$ immediately precedes $\mathbf{x}'$ in the order. Because no influence of $X_2$ on $Y$ has been specified, the parent configurations that have a different value for $X_2$ are incomparable. As a consequence, the order consists of two components, one for $X_2 = 0$ and one for $X_2 = 1$. Estimates may be computed for the two components separately, because there are no order constraints between parent configurations contained in different components. Also note that the component for $X_2 = 0$, depicted in the left part of figure 2, contains a cycle as a result of the context-specific independence specified for $X_3$. Because of the independence the constraint $p^*(y = 1|1, 0, 0) = p^*(y = 1|1, 0, 1)$ must be satisfied. This is modelled by considering the two parent configurations $(1, 0, 0)$ and $(1, 0, 1)$ as a single element in the ordering, as shown in figure 3.

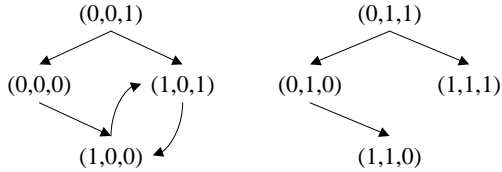

Figure 2: Order corresponding to the network fragment

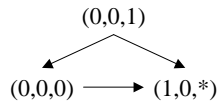

Figure 3: Updated order for $X_2 = 0$

Table 1 shows for each parent configuration with $X_2 = 0$ the counts obtained from a given sample, as well as the associated maximum-likelihood estimates $\hat{p}(y = 1|\mathbf{x})$. We note that the estimates should be non-decreasing in both columns of the table, due to the positive influence of $X_1$ on $Y$. This constraint is violated in the first column. Furthermore, the estimates in the second row should be equal, due to the context-specific independence specified. Clearly, this constraint is violated as well. Finally, the estimates should be non-increasing in the first row, due to the context-specific negative influence of $X_3$. This constraint is satisfied.

The MLS algorithm resolves the identified constraint violations by averaging the unconstrained estimates $\hat{p}(y = 1|\mathbf{x})$ over conflicting cells from the table. It starts with the computation of the weighted average of $\hat{p}(y = 1|\mathbf{x})$ for all lower sets of $\Omega(\mathbf{X})$; table 2 shows the resulting averages. The minimum average is achieved by $\{(0, 0, 1)\}$, so the algorithm sets $p^*(y = 1|0, 0, 1) = 1/5$. This element is removed from all lower sets, and their weighted averages are recomputed. Now $\{(0, 0, 0)\}$ has the minimum weighted average, and we get $p^*(y = 1|0, 0, 0) = 4/10$. The element $(0, 0, 0)$ is removed from all lower sets, and $\{(1, 0, 0), (1, 0, 1)\}$ is the only remaining lower set. Its weighted average is $10/23$, so the algorithm sets $p^*(y = 1|1, 0, 0) = p^*(y = 1|1, 0, 1) = 10/23$. Now the component of the order with $X_2 = 0$ has been solved. Note that the two constraint violations have been resolved simultaneously by averaging the pair of violators $\hat{p}(y = 1|1, 0, 0)$ and $\hat{p}(y = 1|1, 0, 1)$.

Next we consider the parent configurations with $X_2 = 1$. Table 3 shows for each such parent configuration the counts obtained from the available sample, and the associated maximum-likelihood estimates $\hat{p}(y = 1|\mathbf{x})$. Note that there are no observations in the sample with the parent configuration $(0, 1, 1)$. In such cases we put $\hat{p}(y = 1|\mathbf{x}) = 0.5$ and give the cell an arbitrary small weight. As a consequence the estimate will the be dominated by other parameter estimates as soon as it is pooled to resolve conflicts.

From the specified signs, we have that the estimates should be non-decreasing within each column, and non-increasing in the first row. The row constraint is satisfied, but the column constraints are not. Table 4 gives all lower sets with $X_2 = 1$, and their weighted averages. The set $\{(0, 1, 1), (1, 1, 1)\}$ achieves the minimum and the MLS algorithm sets $p^*(y = 1|0, 1, 1) = p^*(y = 1|1, 1, 1) = 0.4$. Note that the constrained estimate for the empty cell $(0, 1, 1)$ has simply been set equal to the estimate for the conflicting cell $(1, 1, 1)$. The elements $(0, 1, 1)$ and $(1, 1, 1)$ are removed from all lower sets, and the weighted averages are recomputed. Now the minimum weighted average of $12/25 = 0.48$ is achieved by $\{(0, 1, 0), (1, 1, 0)\}$, so we get $p^*(y = 1|0, 1, 0) = p^*(y = 1|1, 1, 0) = 0.48$. In this step the violation of the order constraint in the first column is resolved by averaging the parameter estimates for the two conflicting cells. Note that the constrained joint estimate is closer to the unconstrained estimate for cell $(0, 1, 0)$ than to the unconstrained estimate for cell $(1, 1, 0)$ because we have more observations in the former than in the latter and the former thus has a larger associated weight.

Table 1: Counts and ML estimates for $X_2 = 0$

| $X_2 = 0$ | $X_3 = 0$ | $X_3 = 1$ |
|---|---|---|
| $X_1 = 0$ | $4/10 = 0.4$ | $1/5 = 0.2$ |
| $X_1 = 1$ | $6/18 = 0.33$ | $4/5 = 0.8$ |

Table 2: The weighted average of the lower sets for $X_2 = 0$

| Lower set | Weighted average | | |
|---|---|---|---|
| | 1 | 2 | 3 |
| $\{(0,0,1)\}$ | **1/5** | – | – |
| $\{(0,0,1),(0,0,0)\}$ | 5/15 | **4/10** | – |
| $\{(0,0,1),(0,0,0),(1,0,0),(1,0,1)\}$ | 15/38 | 14/33 | **10/23** |

Table 3: Counts and ML estimates for $X_2 = 1$

| $X_2 = 1$ | $X_3 = 0$ | $X_3 = 1$ |
|---|---|---|
| $X_1 = 0$ | $10/20 = 0.5$ | $0/0 = 0.5$ |
| $X_1 = 1$ | $2/5 = 0.4$ | $4/10 = 0.4$ |

Table 4: The weighted average of the lower sets for $X_2 = 1$

| Lower set | Weighted average | |
|---|---|---|
| | 1 | 2 |
| $\{(0,1,1)\}$ | 0.5 | – |
| $\{(0,1,1),(0,1,0)\}$ | 10/20 | 10/20 |
| $\{(0,1,1),(1,1,1)\}$ | **4/10** | – |
| $\{(0,1,1),(0,1,0),(1,1,0)\}$ | 12/25 | **12/25** |
| $\{(0,1,1),(0,1,0),(1,1,1)\}$ | 14/30 | – |
| $\{(0,1,1),(0,1,0),(1,1,0),(1,1,1)\}$ | 16/35 | – |

We would also like to note that, although the algorithm computes $p^*(y = 1|0,1,1) = 0.4$ for the empty cell $(0, 1, 1)$, any value in the interval $[0, 0.4]$ would actually have satisfied the constraints. An alternative to the proposed procedure would be to remove the empty cells before application of the MLS algorithm, and after the estimates for the other cells have been computed, determine feasible estimates for the empty cells.

Since $\Omega(\mathbf{X})$ is exhausted after this step, the algorithm halts. We observe that the resulting parameter estimates indeed satisfy the constraints imposed by the qualitative influences. Also note that the parameters that have not been involved in any constraint violations have retained their original estimates.

### 4.2 COMPLEXITY OF THE MLS ALGORITHM

We argued in section 3 that the number of lower sets is the dominant factor in the runtime complexity of the minimum lower sets algorithm. To determine this number, we start with the simple case where all $k$ parents of a variable have a context-independent signed influence. Without loss of generality, we assume all signs to be positive. Since all parents are binary, any parent configuration from $\Omega(\mathbf{X})$ is uniquely determined by the parents that have the value 1, or alternatively, by a subset of $\{1, 2, \ldots, k\}$. The partial order on $\Omega(\mathbf{X})$ is therefore isomorphic to the partial order generated by set inclusion on $\mathcal{P}(\{1, 2, \ldots, k\})$. Every lower set now corresponds uniquely to an antichain in this partial order. Hence, the number of distinct lower sets equals the number of distinct nonempty antichains of subsets of a $k$-set, which adheres to the well-known Sloane sequence A014466. Writing $|\mathcal{L}(k)|$ for the number of lower sets for $k$ parents as above, we thus find that $|\mathcal{L}(5)| = 7580$, $|\mathcal{L}(6)| = 7828353$, and $|\mathcal{L}(7)| = 2414682040997$. We conclude that the MLS algorithm is feasible for up to five or six parents with signed influences only.

From our example network fragment, we noted that unsigned influences serve to partition the set of parent configurations $\Omega(\mathbf{X})$ into disjoint subsets, such that no element of the one subset is order related to any element of the other subsets. We argued that constrained estimates may be computed for these subsets separately, thereby effectively decomposing the parameter learning problem into a number of independent smaller problems. This decomposition yields a considerable improvement of the efficiency of the computations involved. In general, let $k_1$ denote the number of parents with a signed influence and let $k_2$ denote the number of parents with an unsigned influence. The number of configurations for the parents with an unsigned influence equals $2^{k_2}$. The order graph thus consists of $2^{k_2}$ components. The number of lower sets of the entire order is given by

$$|\mathcal{L}(k_1 + k_2)| = (|\mathcal{L}(k_1)| + 1)^{2^{k_2}} - 1$$

This number grows very rapidly. For $k_1 = 4$ and $k_2 = 3$, for example, the algorithm would need to compute the weighted average of $168^8 - 1 = 6.35 \times 10^{17}$ lower sets. By treating each component in the order as a separate problem, the algorithm initially has to compute the weighted average of

$$|\mathcal{L}(k_1 + k_2)| = 2^{k_2} |\mathcal{L}(k_1)|$$

lower sets. For $k_1 = 4$ and $k_2 = 3$, this amounts to just $8 \cdot 167 = 1336$ lower sets.

In the presence of context-specific signs, the analysis of the algorithm's runtime complexity becomes more complicated. We restrict ourselves to the following observations. First, the absence of signs in particular contexts can also lead to a decomposition of the order, and hence to a similar reduction of the computations involved as in the case of unsigned influences. On the other hand, the absence of signs in particular contexts can also lead to an increase of the number of lower sets. Secondly, context-specific independences lead to a reduction of the number of elements in the ordering, that is of the number of parameters to be estimated, and hence to a reduction of the number of lower sets.

## 4.3 BAYESIAN ESTIMATION

The parameter learning method described in the previous sections does not require that an expert specifies numerical values for the parameters concerned. The expert only has to provide signs for the various influences. Should uncertain prior knowledge about the numeric values of the parameters be available in addition to knowledge about the signs of influences, then we can accommodate this information. Suppose that the expert is willing to specify a Beta prior for the parameters $p(y = 1|\mathbf{x}), \mathbf{x} \in \Omega(\mathbf{x})$. We assume that he chooses the hyperparameters $a(\mathbf{x})$ and $b(\mathbf{x})$ such that his experience is equivalent to having seen the value $y = 1$ a total of $a(\mathbf{x}) - 1$ times in $h(\mathbf{x}) = a(\mathbf{x}) + b(\mathbf{x}) - 2$ trials; $h$ is called the prior precision. Let $p_0(y = 1|\mathbf{x})$ denote the modal value of $p(y = 1|\mathbf{x})$, that is, $p_0(y = 1|\mathbf{x})$ is a priori considered the most likely value of $p(y = 1|\mathbf{x})$. We now further assume that the expert's values for $a(\mathbf{x})$ and $b(\mathbf{x})$ are such that the modes $p_0(y = 1|\mathbf{x}) = (a(\mathbf{x}) - 1)/h(\mathbf{x}), \mathbf{x} \in \Omega(\mathbf{X})$, are isotonic with respect to the order imposed by the signs he specified.

In forming the joint prior for $p(y = 1|\mathbf{x}), \mathbf{x} \in \Omega(\mathbf{x})$, we assume local parameter independence (cf. [10]), except that the parameter values must be isotonic. This means that the prior density is 0 for non-isotonic value combinations for the parameters, and proportional to the product Beta distribution for isotonic value combinations. The isotonic MAP estimates then are given by the isotonic regression of

$$p_0(y = 1|\mathbf{x}, \mathcal{D}) = \frac{n(\mathbf{x})\hat{p}(y = 1|\mathbf{x}) + h(\mathbf{x})p_0(y = 1|\mathbf{x})}{n(\mathbf{x}) + h(\mathbf{x})}$$

with weights $n(\mathbf{x}) + h(\mathbf{x})$ (see [1]).

As before order-constrained estimation now amounts to performing isotonic regression on basic estimates with appropriately chosen weights. The basic estimates are the unconstrained MAP estimates $p_0(y = 1|\mathbf{x}, \mathcal{D})$ for the parameters. The weight is $n(\mathbf{x}) + h(\mathbf{x})$, that is the sum of the number of actual observations for parent configuration $\mathbf{x}$ and the prior precision $h(\mathbf{x})$ specified by the expert. Note that in case of a flat prior (Beta(1,1); $h = 0$), the order-constrained maximum likelihood estimates are returned.

## 5 EXPERIMENTAL RESULTS

To study the behaviour of the isotonic estimator in a slightly more involved setting, we compare it to the standard maximum-likelihood estimator on the well-known Brain Tumour network [6]; the network and the signs of the influences are depicted in figure 4. For the network, we specified probabilities consistent with the constraints to generate data samples for our experiments. Note that, even though the true distribution satisfies the constraints, this need not hold for the relative frequencies in the samples, especially not for smaller sample sizes.

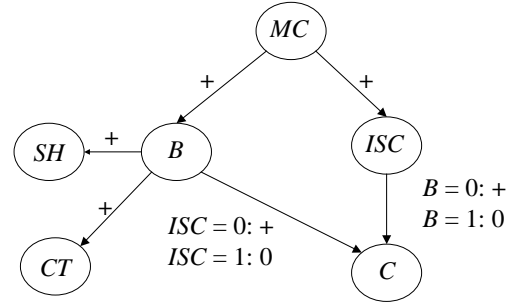

Figure 4: The *Brain Tumour* network

Our implementation first generates, for each variable, the quasi-order on its parent configurations that corresponds with the specified signs. For each order it finds the separate components; all parent configurations contained in the same cycle are mapped to a single element and the order is adjusted accordingly. For each component of the order, the parameters for the parent configurations contained in that component are estimated using the MLS algorithm.

We drew samples of various sizes from the network using logic sampling; for each sample size, 100 data sets were drawn. From each data set, both the standard maximum-likelihood estimates and the constrained estimates of the network parameters were calculated. Given these parameter estimates, the joint distribution defined by the resulting network was computed. This distribution then was compared to the true joint distribution defined by the original network. For comparing the distributions, we used the well-known Kullback-Leibler divergence. The Kullback-Leibler divergence of $\Pr'$ from $\Pr$ is defined as

$$\mathrm{KL}(\Pr, \Pr') = \sum_{\mathbf{x}} \Pr(\mathbf{x}) \log \frac{\Pr(\mathbf{x})}{\Pr'(\mathbf{x})}$$

where a term in the sum is taken to be 0 if $\Pr(\mathbf{x}) = 0$, and infinity whenever $\Pr'(\mathbf{x}) = 0$ and $\Pr(\mathbf{x}) > 0$.

The results are summarized in table 5, where $\widehat{\Pr}$ is used to denote the joint distribution that was obtained with unconstrained maximum-likelihood estimation. To illustrate the benefits of modeling context-specific independences, we first estimated the various parameters without taking the embedded zeroes into account, that is, we used $B \xrightarrow{+} C$, and $ISC \xrightarrow{+} C$. The resulting distribution is denoted by $\Pr^*$ in the table. Finally, $\Pr^{**}$ denotes the distribution that was obtained with the isotone estimator using the embedded zeroes. The averages reported in the table were computed from those data sets for which the KL divergence was smaller than infinity for the maximum-likelihood estimates: the isotone estimates always have KL divergence smaller than infinity in these cases as well. The number of data sets from which the averages were computed was 61, 97, 100, and 100 for sample sizes 50, 150, 500, and 1500, respectively. For sample sizes 20, 30, and 40, we used Bayesian estimation with a Beta(2,2) prior for all parameters, that is, the prior mode of all parameters was set to 0.5 with prior precision equal to 2. Note that by setting all parameters to the same value we never violate any order constraints. The unconstrained and isotonic MAP estimates were used as point estimates for the parameters. We used Bayesian estimation for the smallest sample sizes because otherwise the KL divergence would almost always be equal to infinity.

Table 5: Experimental results: the Brain Tumour network

| $n$ | $\mathrm{KL}(\mathrm{Pr},\widehat{\mathrm{Pr}})$ | $\mathrm{KL}(\mathrm{Pr},\mathrm{Pr}^*)$ | $\mathrm{KL}(\mathrm{Pr},\mathrm{Pr}^{**})$ |
|---|---|---|---|
| 20 | 0.2149 | 0.1891 | 0.1814 |
| 30 | 0.1572 | 0.1401 | 0.1317 |
| 40 | 0.1442 | 0.1230 | 0.1149 |
| 50 | 0.1286 | 0.1162 | 0.1066 |
| 150 | 0.0481 | 0.0442 | 0.0400 |
| 500 | 0.0132 | 0.0123 | 0.0115 |
| 1500 | 0.0043 | 0.0041 | 0.0036 |

Table 6: Experimental results: the OESOCA network

| $n$ | $\mathrm{KL}(\mathrm{Pr},\widehat{\mathrm{Pr}})$ | $\mathrm{KL}(\mathrm{Pr},\mathrm{Pr}^*)$ |
|---|---|---|
| 50 | 0.5247 | 0.5044 |
| 100 | 0.2908 | 0.2774 |
| 150 | 0.2005 | 0.1919 |
| 500 | 0.0665 | 0.0640 |
| 1000 | 0.0342 | 0.0333 |
| 1500 | 0.0225 | 0.0219 |

The results reveal that the isotonic estimator consistently yields a better fit of the true distribution compared to the unconstrained maximum-likelihood estimator, although the differences are small. We note that for the smaller data sets the differences are more marked than for the larger data sets. This conforms to our expectations, since for smaller data sets standard maximum-likelihood estimation has a higher probability of yielding estimates that violate the specified constraints. For larger data sets, the standard estimator and the isotonic estimator are expected to often result in the same estimates. Note that using the context-specific independences gives an additional improvement of fit, as was to be expected.

To conclude, we applied the isotonic estimator to a real-life Bayesian network in the field of oncology. The OESOCA network provides for establishing the stage of a patient's oesophageal cancer, based upon the results of a number of diagnostic tests. The network was constructed with the help of gastroenterologists from the Netherlands Cancer Institute, Antoni van Leeuwenhoekhuis; the experts provided the knowledge for the configuration of the network's structure and also provided probability assessments for the network's parameters. From the original OESOCA network, we constructed a binary network for our experiment, carefully building upon knowledge of the domain. The resulting network includes 40 variables with a total of 95 parameters. From values of the the various parameters, we established the signs for the qualitative influences between the variables. The network includes 45 influences; figure 5 depicts the binary OESOCA network. The signs of the qualitative influences are shown over the corresponding arcs; for readability, only the context-independent signs are shown, where a question mark is used to denote an ambiguous influence.

The experimental results are displayed in table 6: $\widehat{\mathrm{Pr}}$ again denotes the joint distribution resulting from the unconstrained MAP estimates, and $\mathrm{Pr}^*$ the joint distribution resulting from the isotonic MAP estimates. The results are in line with the results obtained for the brain tumour network: the isotonic estimator is consistently better, and the difference becomes smaller as the sample size increases.

## 6 CONCLUSIONS

Taking prior knowledge about the signs of influences between variables into account upon estimating the parameters of a Bayesian network, results in an improved fit of the true distribution. The improvement is relatively large for small samples, since these are more likely to give rise to maximum-likelihood estimates that violate the constraints. Since the constrained parameter estimates are in accordance with prior knowledge specified by experts, the resulting network is more likely to be accepted in its domain of application.

An interesting extension of our method would be to allow for non-binary variables with linearly-ordered discrete values. A signed influence on such a variable is defined in terms of a stochastic order on the distributions given the different parent configurations. Learning the parameters of a network subject to the resulting constraints in our opinion merits further research.

### Acknowledgement

The authors would like to thank Kim H. Liem for writing part of the program code, and performing the experiments.

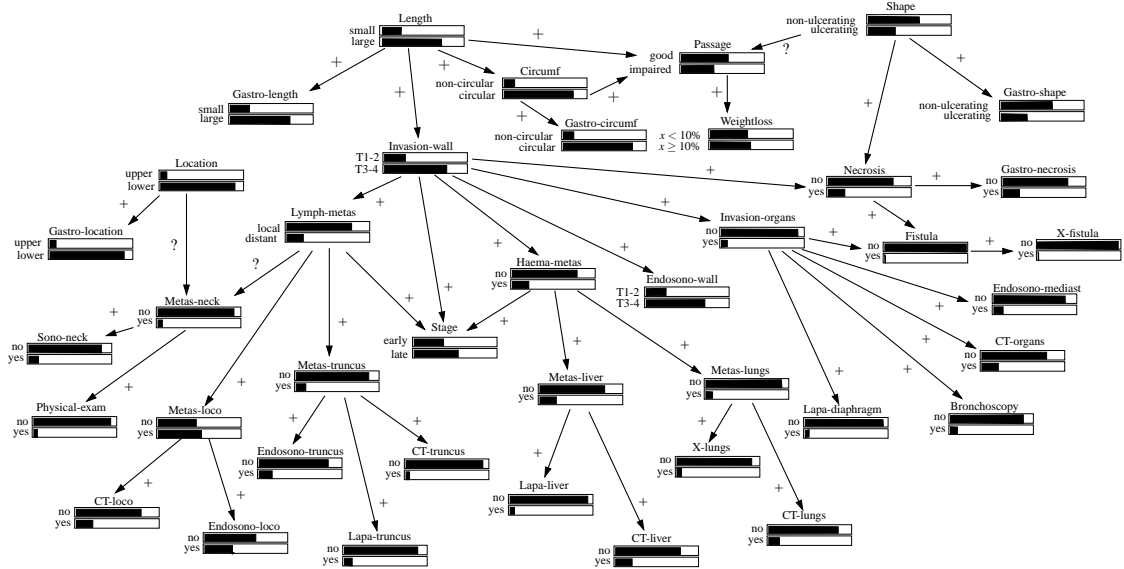

Figure 5: The qualitative OESOCA network